%% file: main.tex
\documentclass[final]{nesy2026} %

\usepackage{longtable}%

\usepackage{booktabs}
\definecolor{dkgreen}{RGB}{0,180,0} %
\usepackage{multirow}
\usepackage{amssymb}
\usepackage{rotating} %
\usepackage[load-configurations=version-1]{siunitx} %
\usepackage{tcolorbox}
\theorembodyfont{\upshape}
\theoremheaderfont{\scshape}
\theorempostheader{:}
\theoremsep{\newline}

\title[Agentic DomiKnowS]{An Agentic Framework for Neuro-Symbolic Programming}

\clearauthor{\Name{Aliakbar Nafar}, \Name{Chetan Chigurupati}, \Name{Danial Kamali},\\
 \Name{Hamid Karimian}, \Name{Parisa Kordjamshidi}\\
 \addr Michigan State University\\
 Correspondence: \href{mailto:nafarali@msu.edu}{nafarali@msu.edu}}

\begin{document}

\maketitle

\input{Content/AbstractIntroduction}

\input{Content/Background}
\input{Content/SystemOverview}
\input{Content/UIImplementation}

\input{Content/Experiments}

\input{Content/RelatedWork}

\input{Content/Limitations}

\input{Content/Conclusion}

\acks{This project is partially supported by the Office of Naval Research (ONR) grant N00014-23-1-2417. Any opinions, findings, and conclusions or recommendations expressed in this material are those of the authors and do not necessarily reflect the views of Office of Naval Research. We thank the anonymous reviewers for their valuable feedback and suggestions, which have helped improve this work.}

\input{main.bbl}
\appendix
\input{Content/Appendix/SelectedTasks}
\input{Content/Appendix/HumanTest}
\input{Content/Appendix/FullResults}
\input{Content/Appendix/CoderPrompts}

\end{document}

%% file: Content/AbstractIntroduction.tex
\begin{abstract}
Integrating symbolic constraints into deep learning models could make them more robust, interpretable, and data-efficient. Still, it remains a time-consuming and challenging task. Existing frameworks like DomiKnowS help this integration by providing a high-level declarative programming interface, but they still assume the user is proficient with the library's specific syntax. We propose AgenticDomiKnowS (ADS) to eliminate this dependency. ADS translates free-form task descriptions into a complete DomiKnowS program using an agentic workflow that creates and tests each DomiKnowS component separately. The workflow supports optional human-in-the-loop intervention, enabling users familiar with DomiKnowS to refine intermediate outputs. We show how ADS enables experienced DomiKnowS users and non-users alike to construct complete neuro-symbolic programs in 10--15 minutes, whereas manually coding even a component of DomiKnowS takes an hour. Access the UI at \url{https://hlr-demo.egr.msu.edu/}.

\end{abstract}

\section{Introduction}
\label{sec:intro}

Deep learning models augmented with symbolic reasoning methods, often referred to as neuro-symbolic (NeSy) systems, combine the strengths of deep learning with the logically consistent symbolic formalisms~\citep{feldstein2024mappingneurosymbolicailandscape}. NeSy systems could make the models more robust, interpretable, and data-efficient than purely neural approaches~\citep{NeuroSymbolicSurveryRelevant}. However, authoring NeSy programs remains challenging, as existing frameworks impose a steep learning curve, requiring users to know the syntax and semantics of the target formalism for each symbolic system they employ~\citep{sinha2025neurosymbolicframeworksconceptualcharacterization}.

\begin{figure}[h!]
    \centering
    \includegraphics[width=1\linewidth]{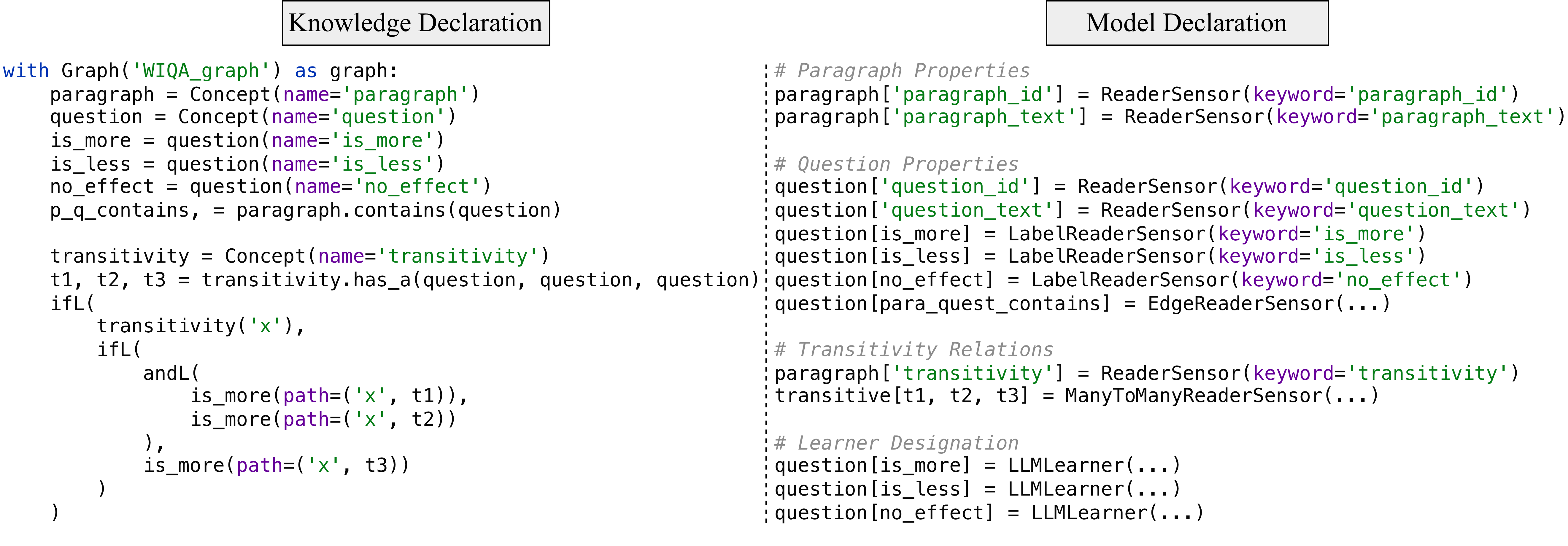}
    \caption{Knowledge and model declarations of a DomiKnowS program for a procedural question-answering task (WIQA dataset). Left side defines the \textit{conceptual graph} and logical constraints, while the right side shows the sensor code, which specifies how properties and predictive models are attached to the graph's concepts.}
    \label{fig:domiknowsoverview}
\end{figure}

NeSy frameworks like DomiKnowS~\citep{rajaby-faghihi-etal-2021-domiknows} exemplify this complexity. As a declarative Python library, DomiKnowS integrates symbolic logic with deep learning models through a user-defined \textit{conceptual graph} encoding concepts, relations, and logical constraints. Consider WIQA~\citep{tandon-etal-2019-wiqa}, a procedural question-answering task. Given a process description, such as the water cycle, the model must predict how perturbing one step affects another. Predictions for these questions fall into three categories: \textit{is\_more}, \textit{is\_less}, or \textit{no\_effect}. Crucially, these answers must obey a \textit{transitivity} constraint. For instance, if increased evaporation ($A$) causes more rain ($B$), and more rain causes flooding ($C$), then $A$ must also increase $C$. Encoding this logic in DomiKnowS requires declaring the relevant concepts, such as the paragraph text and the specific questions, and expressing the transitivity rule using the library's first-order logic syntax as shown in Figure~\ref{fig:domiknowsoverview}. While the \textit{conceptual graph} offers significant flexibility, authoring the program is demanding. It requires mastering the library's syntax and manually encoding every rule, a time-consuming and error-prone process for anyone without prior experience.

A previous attempt to generate DomiKnowS programs with LLM assistance~\citep{PreviousDomiKnowSUI} produces only the \textit{conceptual graph} and its constraints, and requires significant intervention by users already familiar with the library. While newer LLMs have improved at expressing logical constraints and high-level relational structures~\citep{ijcai2025p1155,shi-etal-2025-constraintllm,lalwani2025autoformalizingnaturallanguagefirstorder}, they continue to struggle with mapping them into correct syntax for domain-specific libraries. Additionally, because NeSy libraries like DomiKnowS are underrepresented or non-existent in pre-training corpora, LLMs frequently fail to generate their programs~\citep{10.1145/3697012,abbassi2025taxonomyinefficienciesllmgeneratedpython}.

In this work, we introduce our workflow to overcome these generation barriers and enable users to create NeSy programs using natural language instructions and task descriptions. Given only a natural-language task description, ADS produces a complete and executable DomiKnowS program, including the \textit{conceptual graph}, its logical constraints, and the learners attached to it, without requiring the user to write any DomiKnowS-specific syntax. Unlike coding assistants such as GitHub Copilot~\citep{github_copilot} or Cursor~\citep{cursor}\footnote{These assistants proved incapable of generating DomiKnowS programs from documentation. We show the failure modes of these assistants in our experiments.} that synthesize an entire program at once, ADS employs an agentic workflow that breaks the development process into distinct stages, generating, executing, and refining each code section independently. This allows ADS to isolate and fix, in an iterative self-refinement loop~\citep{madaan2023selfrefine}, syntactic errors identified during code execution and semantic errors detected by an LLM-based reviewer. ADS supports optional human-in-the-loop intervention through an interactive user interface (UI) that visualizes the information needed for verification or refinement. The interface generates the DomiKnowS program as a plug-and-play executable Jupyter notebook, utilizing general-purpose Vision-Language Models (VLMs) as pre-initialized learning models for immediate inference.

In summary, our main contributions are as follows: 1) We introduce AgenticDomiKnowS~\footnote{The code is available at \url{https://github.com/HLR/AgenticDomiKnowS} and the UI at \url{https://hlr-demo.egr.msu.edu/}.}, an agentic workflow with an interactive human-in-the-loop UI that translates natural language task descriptions into NeSy programs. 2) We demonstrate that ADS enables both DomiKnowS experts and non-users to construct complete NeSy programs in 10--15 minutes, compared to one hour for manually authoring just one component of it~\citep{PreviousDomiKnowSUI}. 3) We showcase various limitations of current state-of-the-art coding assistants in generating NeSy code for low-resource libraries such as DomiKnowS and demonstrate how our agentic approach overcomes these barriers.

%% file: Content/Background.tex
\section{Background: DomiKnowS}

A DomiKnowS program is written in a domain-specific, Python-based language and is organized into two parts: a \textit{Knowledge Declaration} and a \textit{Model Declaration}. The Knowledge Declaration specifies a \textit{conceptual graph}, namely the concepts of the domain, the relations among them, and the first-order logical constraints expressed over these concepts using operators such as \texttt{ifL} and \texttt{andL}. The Model Declaration attaches data and trainable models to the graph through predefined \textit{sensors} and learners, after which the program can be executed to produce predictions that satisfy the declared constraints. We illustrate both parts using the WIQA running example introduced in Section~\ref{sec:intro} and depicted in Figure~\ref{fig:domiknowsoverview}.

For WIQA, the Knowledge Declaration creates a \textit{paragraph} concept and a \textit{question} concept. We define the answer classes \textit{is\_more}, \textit{is\_less}, and \textit{no\_effect}, which serve as labels for the questions. We define a one-to-many containment relation between paragraphs and questions. To express the relations between questions, we introduce a \textit{transitivity} concept that connects each triplet of questions ($t_1, t_2, t_3$), where $t_1$ and $t_2$ ask about consecutive links of a causal chain (the effect of $A$ on $B$ and of $B$ on $C$) and $t_3$ asks about its endpoints (the effect of $A$ on $C$). Finally, we add a logical constraint using the \texttt{ifL} and \texttt{andL} operators: whenever $t_1$ and $t_2$ are both answered \textit{is\_more}, the constraint forces $t_3$ to also be \textit{is\_more}.

In the Model Declaration, we load the data elements from the dataset and attach them as properties to the corresponding concepts using predefined DomiKnowS sensors (e.g., \texttt{ReaderSensor}). We also assign trainable models here: an \texttt{LLMLearner} that wraps a Vision-Language Model to predict question labels. We populate the graph's relations, such as the transitivity triplets provided by the dataset. After these steps, the program can be executed to infer the output labels while obeying the logical constraints using Integer Linear Programming (ILP)~\citep{ILP}. Moreover, we can train the models to learn to satisfy the constraints using the underlying algorithms proposed in related research~\citep{Lee_Mehta_Wick_Tristan_Carbonell_2019,NEURIPS2019_cf708fc1,ijcai2020p382}.

%% file: Content/SystemOverview.tex
\section{System Overview}

\begin{figure}
    \centering
    \includegraphics[width=1.0\linewidth]{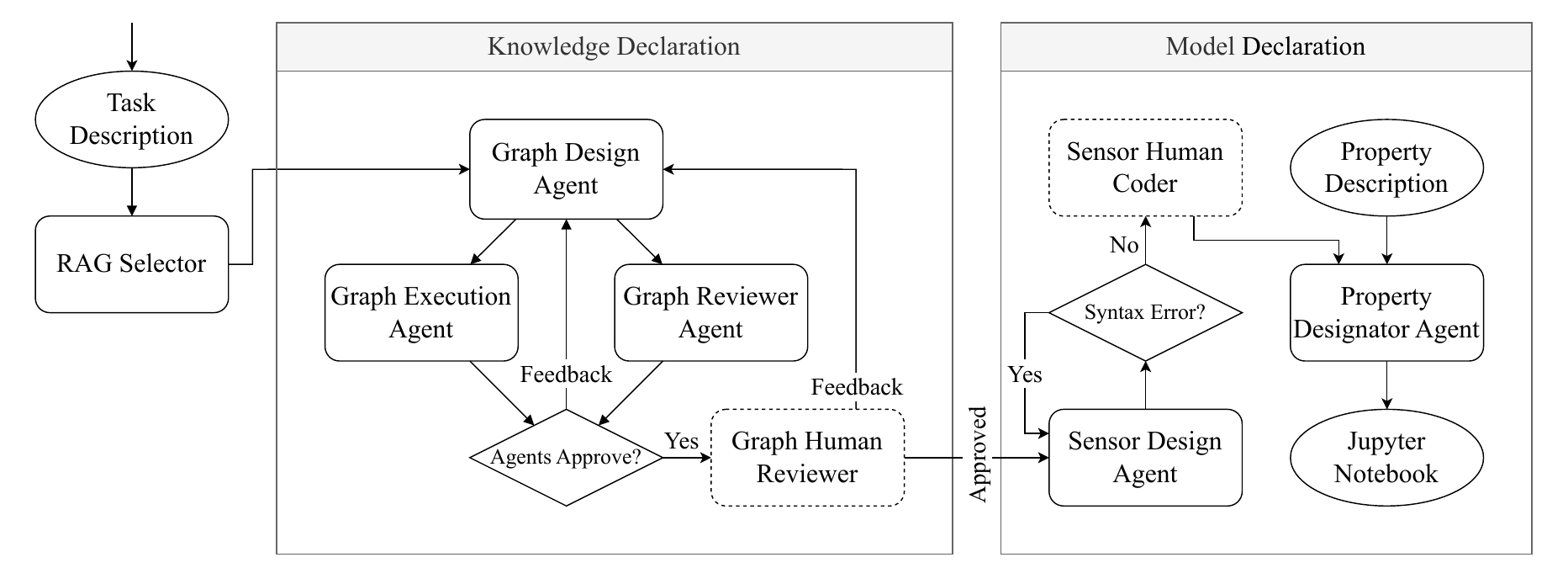}
    \caption{ADS system workflow. The left panel groups components of the \textit{Knowledge Declaration} phase and the right panel the \textit{Model Declaration} phase. Rectangles denote LLM agents and dashed rectangles denote human actions.}
    \label{fig:ADSworkflow}
\end{figure}

ADS is implemented as an agentic workflow over a shared memory state that stores the task description, code drafts and reviews. As depicted in Figure~\ref{fig:ADSworkflow}, the workflow begins with a retrieval step in which, given the input task description, the \textit{RAG Selector} agent searches a pool of prior DomiKnowS programs, which are authored by users of the DomiKnowS library, and retrieves the most relevant ones based on their task descriptions. For the WIQA task, for example, it retrieves RuleTaker~\citep{RuleTaker}, a deductive reasoning task that chains natural-language if-then rules, closely mirroring WIQA's transitivity constraint. These serve as in-context examples for the LLMs later in the workflow. Then ADS proceeds through the two main phases of Knowledge Declaration and Model Declaration.

\subsection{Knowledge Declaration}

The next step is generating the Knowledge Declaration code, which produces a graph code comprising the \textit{conceptual graph} and first-order logical constraints. Knowledge Declaration is handled by a feedback loop between three agents: The \textit{Graph Design Agent} proposes a new graph code implementation based on the task description and the retrieved RAG examples. Subsequently, the \textit{Graph Execution Agent} executes the newly generated graph code and records the output errors. For WIQA, it commonly flags syntactic mistakes such as accessing a triplet element through an invalid attribute like \texttt{transitivity.t1}. In parallel, the \textit{Graph Reviewer Agent} produces a natural-language review of the generated graph to verify its semantics. This reviewer suggests that the constraints be modified if needed, relation definitions corrected, and unnecessary concepts removed. In WIQA, it often catches the transitivity constraint being declared two or three times over different permutations of the triplet $(t_1, t_2, t_3)$. If the generated graph is not approved by either or both the \textit{Graph Execution Agent} and the \textit{Graph Reviewer Agent}, it will be sent back to the \textit{Graph Design Agent} along with the feedback.

When the agents either approve a draft or hit the attempt limit, the workflow reaches the \textit{Graph Human Reviewer}, which asks the user whether they approve the graph and collects feedback otherwise. If the user rejects the graph, the system clears accumulated review and execution notes, resets the attempt counter, and sends control back to the \textit{Graph Design Agent} for a new round of revisions while applying the human feedback. If the user approves, the controller advances to the next stage.

\subsection{Model Declaration}

In the Model Declaration stage, ADS assigns sensors and learners to the graph concepts. A central design decision in ADS is to minimize the use of some complex DomiKnowS classes without losing any DomiKnowS functionalities and instead delegating their operations to ordinary Python code, which LLMs are generally better at generating. This results in sensors being the i) \textit{ReaderSensor}, which is used to read the data elements, ii) \textit{LabelReaderSensor}, which is used to read the labels/annotations, iii) \textit{EdgeReaderSensor}, which connects the concepts in a containment relation, and iv) \textit{ManyToManyReaderSensor}, which connects the concepts in a \textit{has\_a} relationship. The Learner Module in DomiKnowS, which assigns a deep learning model to predict a label, is replaced by a multi-purpose VLM named \textit{LLMLearner}. This allows the learner to process multimodal inputs. The learner can perform zero-shot inference or be fine-tuned within the DomiKnowS framework for improved accuracy.

The \textit{Sensor Design Agent} generates the sensor code based on the task description, generated graph, and the RAG retrieved sensor codes. If syntax errors occur, it attempts one automatic refinement. The resulting code is then handed to the \textit{Sensor Human Coder}, who can either approve it or edit it further. In the second step of the Model Declaration stage, the user explains in natural language how dataset elements map to the graph concepts. This assignment is deferred to this stage to accommodate varying graph designs. Using the user's description and retrieved RAG examples, the \textit{Property Designator Agent} updates the code to bind dataset elements as concept properties. Additionally, the agent generates prompts for the VLMs, specifying their outputs. In the exported program, VLM predictions are reconciled with the constraints, so that for WIQA the answers obey transitivity, which can be enforced at inference time or learned by the models during training, both supported in DomiKnowS. The final code is exported to a Jupyter notebook containing cells for the graph, dataset, and sensor sections, along with the DomiKnowS installation commands.

%% file: Content/UIImplementation.tex
\section{Human-in-the-Loop Interface}

\begin{figure*}
    \centering
    \includegraphics[width=1.0\linewidth]{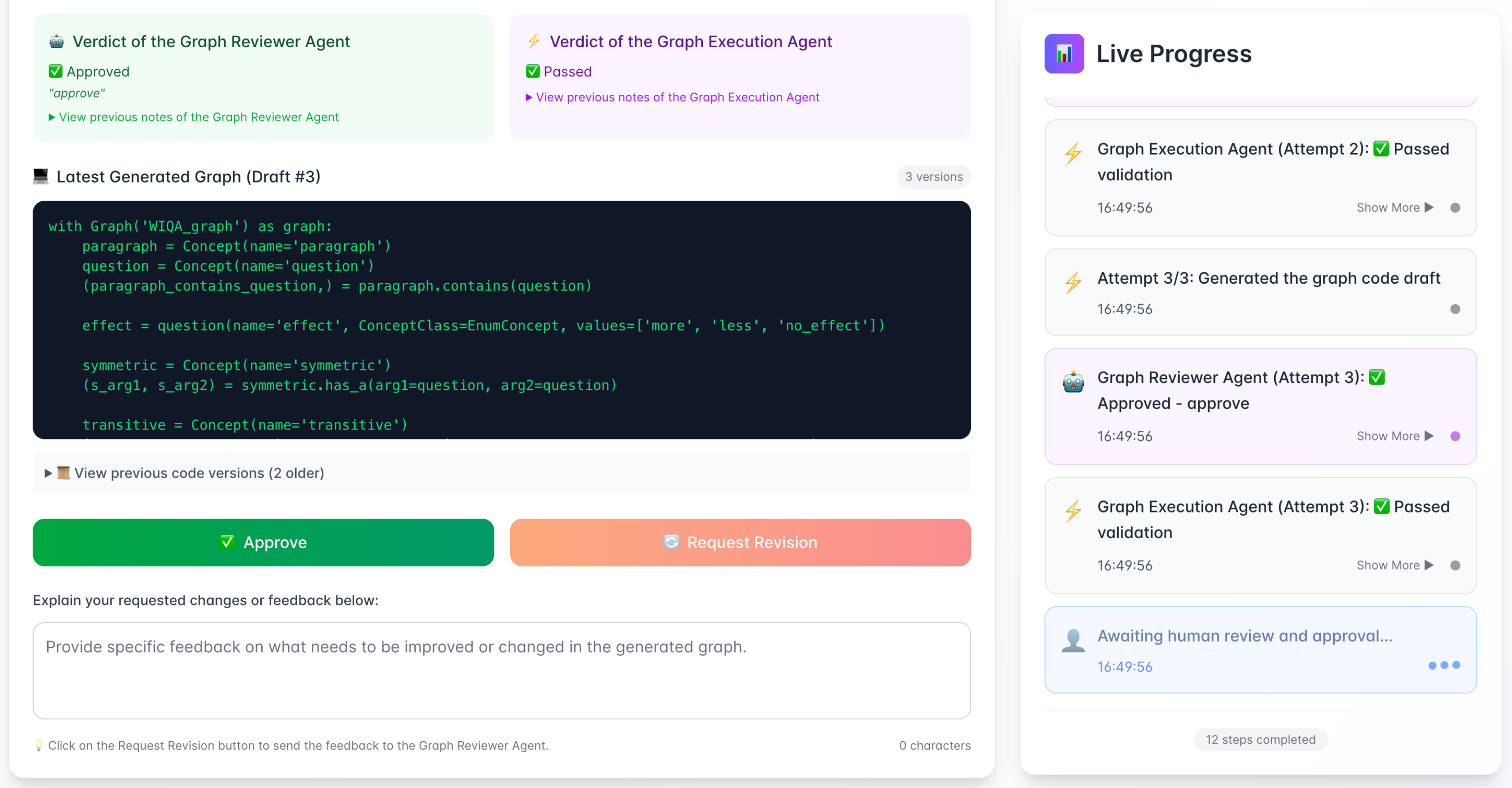}
    \caption{The \textit{Graph Human Reviewer} Interface. The dashboard presents the latest generated graph draft (left) alongside the verdicts from the Reviewer and Execution agents. The user can track the agent workflow history in the ``Live Progress'' panel (right) and either approve the code or provide feedback for revision.}
\label{fig:human_feedback_ui}
\end{figure*}

One of ADS's key features is its human-in-the-loop capability, enabled by a UI that allows anyone, including participants in our testing, to interact directly with the system. %
At each step of the workflow shown in Figure~\ref{fig:ADSworkflow}, the interface retrieves the complete LangGraph memory state but dynamically renders only the information relevant to the current specific stage. For example, as illustrated in Figure~\ref{fig:human_feedback_ui}, the \textit{Graph Human Reviewer} is presented with a view containing the latest generated graph code, agent verdicts, and execution logs. This design ensures that the user has access to all context necessary to make an informed decision. A DomiKnowS-familiar user reviewing the WIQA graph can, for instance, request structural changes in natural language, such as using a single \textit{paragraph\_question} concept instead of separate \textit{paragraph} and \textit{question} concepts for simplicity (both semantically correct). The feedback is routed back to the \textit{Graph Design Agent}, which regenerates the graph accordingly.

%% file: Content/Experiments.tex
\section{Experiments}

We evaluate the two core stages of the ADS workflow, including an ablation isolating the contributions of the RAG and reviewer agents, then present a user study with DomiKnowS experts and non-users to quantify the reduction in development time. Finally, we benchmark ADS against general-purpose coding assistants on the same tasks as the user study.

\noindent\textbf{Datasets.} We use a test dataset of 12 DomiKnowS programs that span NLP, Computer Vision, and Constraint Satisfaction Problems (CSP). The NLP tasks encompass hierarchical topic classification on the 20 News dataset~\citep{Lang95}, logical and causal reasoning covering RuleTaker~\citep{RuleTaker}, WIQA~\citep{tandon-etal-2019-wiqa}, BeliefBank~\citep{kassner-etal-2021-beliefbank}, and ProPara~\citep{dalvi-etal-2018-tracking}, as well as consistency-constrained classification using Spam~\citep{EmailSpam} and IMDB~\citep{maas-etal-2011-learning}. In the Vision domain, we utilize hierarchical image recognition with hierarchically grouped CIFAR-10~\citep{CIFAR10} and Animals \& Flowers image classification~\citep{animalflowers} alongside arithmetic constraint tasks like MNIST-Sum~\citep{DeepProbLog}. Finally, the CSP category includes classical combinatorial inference problems such as Sudoku and Eight Queens to test strict logical constraint satisfaction. Refer to the Appendix~\ref{appendix:selectedtasks} for more details about these tasks. Beyond serving as our test set, these 12 user-authored programs also form the retrieval corpus for ADS's RAG component. To prevent data leakage, when we evaluate a given test task we exclude its own program from the corpus in a leave-one-out setup, so the workflow never retrieves the target. For the human evaluation, we introduce three additional tasks that are not part of the 12, ordered by increasing constraint complexity. Amazon Reviews~\citep{keung-etal-2020-multilingual} is an unconstrained task that predicts a 1--5 star rating for each review. Hierarchical classification of scientific papers, WOS~\citep{WOS}, predicts a coarse and a fine category per article, subject to the constraint that the fine label must agree with its coarse parent. Sequence labeling of the CoNLL~\citep{tjongkimsang2003conll} dataset assigns an IOB tag to each token, subject to constraints on tag transitions, e.g., an ``I'' tag cannot immediately follow an ``O'' tag.

\noindent\textbf{Language Models.} For our experiments, we employ GPT-5~\citep{openai2025gpt5} version gpt-5-2025-08-07 to evaluate performance across reasoning levels ranging from minimal to medium. We also used two open-weight models: Kimi-K2-Thinking~\citep{kimik2} and DeepSeek-R1-0528~\citep{deepseek2025r1}. 

\subsection{Evaluation of ADS Workflow}

The first significant step of our workflow is Knowledge Declaration, where the \textit{conceptual graph} is generated. To evaluate a generated graph, we compare it against a \emph{reference graph}: the original, human-authored DomiKnowS program for that task, which encodes the intended concepts, relations, and constraints. We sample each task in our test dataset three times, for a total of 36 graphs, and two annotators label each graph into one of three categories. A graph is \textbf{Minimal Valid} if it is semantically equivalent to the reference graph while using only the necessary clauses, \textbf{Redundant Valid} if it is semantically equivalent but adds extra yet harmless clauses (e.g., a repeated logical constraint), and \textbf{Semantically Incorrect} if it executes but is not semantically equivalent to the reference graph. We consider a graph \emph{valid} when it is either Minimal or Redundant Valid, and report this combined rate as \textbf{Valid}. As shown in Table~\ref{tab:model_percentages_updated}, the open-weight models DeepSeek R1 and Kimi k2 achieved the best results, with Kimi k2 reaching 97.22\% validity. However, their inference latency is too high for an interactive UI, so we use GPT-5 (Low) in this stage for its superior speed--accuracy trade-off. To see the detailed results, refer to the Appendix~\ref{FullResults}.

\begin{table}[h!]
    \centering
    \small
    \renewcommand{\arraystretch}{1.2}
    \setlength{\tabcolsep}{4.5pt}
    \begin{tabular}{@{}l ccc cc@{}}
    \toprule
     & \multicolumn{3}{c}{\textbf{Full System}} & \multicolumn{2}{c}{\textbf{Ablation}} \\
    \cmidrule(lr){2-4} \cmidrule(lr){5-6}
    \textbf{Model} & \textbf{Minimal Valid} & \textbf{Redundant Valid} & \textbf{Valid} & \textbf{No RAG} & \textbf{No Reviewers} \\
    \midrule
    GPT-5 (Minimal) & 44.44 & 25.00 & 69.44 & 66.67 & 69.44 \\
    GPT-5 (Low)     & 61.11 & 25.00 & 86.11 & 80.56 & 61.11 \\
    GPT-5 (Medium)  & 69.44 & 13.89 & 83.33 & 75.00 & 69.44 \\
    DeepSeek R1     & 69.44 & 19.44 & 88.89 & 47.22 & 77.78 \\
    Kimi k2         & 86.11 & 11.11 & 97.22 & 69.44 & 86.11 \\
    \bottomrule
    \end{tabular}
    \caption{Graph validity per model. The \textit{Full System} columns give the validity breakdown of Minimal Valid, Redundant Valid and Valid graphs. The \textit{Ablation} columns report validity when removing retrieval or the \textit{Graph Reviewer and Execution Agents}.}
    \label{tab:model_percentages_updated}
\end{table}

To isolate the contribution of ADS's multi-agent design, we ablate two components and report total graph validity in the \textit{Ablation} columns of Table~\ref{tab:model_percentages_updated}. In the \textit{No RAG} setting we remove the retrieval of in-context examples, and in the \textit{No Reviewers} setting we remove the \textit{Graph Reviewer and Execution Agents}. Removing either component degrades every model to different degrees. The only exception is GPT-5 (Minimal) under \textit{No Reviewers}, which is already performing poorly. The reviewer agents are most critical for our deployed GPT-5 (Low) setting, where its removal drops total validity by $25.00$ points. Retrieval matters most for the open-weight models, with \emph{No RAG} reducing DeepSeek R1 by $41.67$ points and Kimi k2 by $27.78$ points, while GPT-5 does well with the general-purpose examples.

For the Model Declaration stage, we set the graph generator model to GPT-5 (Low) and vary the reasoning level only for the model-code generation to test the workflow end-to-end. We sample each of the 12 test tasks 5 times, yielding 60 runs in total. GPT-5 with minimal, low, and medium reasoning results in 20, 14, and 11 failures (runs whose generated program does not execute successfully end-to-end), respectively. Because this stage does not employ the large iterative repair loop used in Knowledge Declaration, we choose GPT-5 (Medium) for Model Declaration, as it provides the best accuracy while maintaining acceptable inference time. By manually inspecting three runs per task, we observed that any remaining issues in the code, aside from semantically incorrect graphs, are reliably captured by running the program. Refer to Appendix~\ref{FullResults} for detailed results.

\subsection{Human Evaluation}
\label{sec:humaneval}

We assessed ADS with nine participants (four DomiKnowS experts and five non-users) using three progressively difficult tasks. These difficulties are measured based on the time taken to solve these tasks by manually authoring the code. We measured the development time for each task, excluding runtime. Detailed information regarding the task specifications and participant instructions is available in Appendix~\ref{appendix:humantest}. A prior controlled study found that, even after a tutorial and with full access to documentation and examples, participants took on average about an hour to hand-code just the knowledge declaration of a task~\citep{PreviousDomiKnowSUI}. In contrast, using ADS, DomiKnowS users finish complete programs for most tasks in about 10--15 minutes, as shown in Table~\ref{tab:user_study}.

Users 1 and 3 only edited code to read the dataset inputs. User 2 modified the sensor code for Task 2 and asked the model to revise the graph in Task 3. User 4 also provided graph feedback to modify Task 1 multiple times, which resulted in a longer development time, but finished Tasks 2 and 3 quickly. User 5 completed Tasks 1 and 3 easily, but found Task 2's semantically correct batch definition (article\_group.contains(article)) difficult to resolve. User 6 succeeded without issues, debugging Task 2 despite lacking familiarity with the framework. User 7 also succeeded in all tasks and spent 5 minutes debugging Task 3's code. User 8 spent time resolving the batch in the graph (amazon\_product.contains(reviews)) in Task 1, but had no issues in Tasks 2 and 3. User 9 used graph feedback multiple times for Task 3 and spent time debugging at the end, which ultimately resolved the code.

\begin{table}[h]
\small
    \centering
    \begin{tabular}{l cccc ccccc}
        \toprule
        & \multicolumn{4}{c}{\textbf{Users}} & \multicolumn{5}{c}{\textbf{Non-Users}} \\
        \cmidrule(lr){2-5} \cmidrule(lr){6-10}
         & 1 & 2 & 3 & 4 & 5 & 6 & 7 & 8 & 9 \\ 
        \midrule
        Task 1 & 4 & 6  & 7 & 20 & 7 & 15 & 9 & 18 & 10\\
        Task 2 & 10 & 10  & 9 & 10 & F & 15 & 10 & 15 & 10\\
        Task 3 & 11 & 15  & 20 & 10 & 10 & 8 & 15 & 11 & 25\\
        \bottomrule
    \end{tabular}
    \caption{Development time in minutes by task and user. F indicates failure to complete.}
    \label{tab:user_study}
\end{table}

\subsection{Coding Assistant Evaluation as a Baseline}

Because no prior system synthesizes DomiKnowS programs from a task description, we adopt general-purpose coding assistants, GitHub Copilot~\citep{github_copilot} and Cursor~\citep{cursor}, as the strongest available baseline, evaluated on the three tasks from our human evaluation. Both tools ran the GPT-5.1 Codex Max model (the Mini variant failed on all tasks) with access to the DomiKnowS repository, documentation, and a configured Python environment, and were instructed to write and debug the program without modifying the core library, starting in \textit{plan mode} before transitioning to \textit{agent mode}. See Appendix~\ref{CodingAssistantPrompts} for detailed prompts.

Neither assistant produced a clean, correct program. Cursor's Task 1 code executed but modeled \textit{rating} as a binary concept disconnected from \textit{review} rather than a five-label concept. In Task 2, it defined the coarse and fine labels independently, missing the required hierarchy, and in Task 3, it abandoned the task without producing executable code. Copilot's Task 1 program was the only functionally correct output across both tools, though it improperly imported internal DomiKnowS utility classes to bypass standard implementation constraints. Its Task 2 graph was mostly correct and executable but failed to load the dataset, and, like Cursor, it failed to debug Task 3.

These evaluations highlight three failure modes of general-purpose coding assistants on DomiKnowS programming: 1) They frequently generate syntactically correct but semantically invalid programs, as sound graph designs for novel tasks are absent from the LLMs' training data. 2) They bypass DomiKnowS's loggers and format verifiers by importing internal utility classes and rewriting core logic, producing code that runs but is fundamentally incorrect. 3) They produce bloated, convoluted implementations that are difficult to debug. On the same three tasks, ADS enabled participants, including non-users of DomiKnowS, to produce correct, executable programs in $26$ of $27$ task instances.

%% file: Content/RelatedWork.tex
\section{Related Work}

Integration of neural learning models with symbolic reasoning has led to the development of various NeSy frameworks~\citep{sinha2025neurosymbolicframeworksconceptualcharacterization}. We use DomiKnowS~\citep{rajaby-faghihi-etal-2021-domiknows}, which formulates domain knowledge as a declarative \textit{conceptual graph}, allows switching between various training and inference algorithms, and enables supervision at the granular concept level. For a comprehensive comparison of DomiKnowS and other NeSy frameworks, such as DeepProbLog~\citep{DeepProbLog}, and Scallop~\citep{Scallop}, we refer the reader to~\citet{sinha2025neurosymbolicframeworksconceptualcharacterization}.

The closest prior work is Prompt2DeModel~\citep{PreviousDomiKnowSUI}, an earlier natural-language interface for DomiKnowS, from which ADS departs in three key ways. First, Prompt2DeModel synthesizes only the \textit{Knowledge Declaration}, whereas ADS additionally generates the \textit{Model Declaration} (sensors and learners), producing a complete, executable program. Second, Prompt2DeModel requires the user to validate the graph at every step through a visualization, presuming familiarity with the task structure. In ADS the \textit{Graph Human Reviewer} is an \emph{optional} checkpoint over a draft the agents have already generated and executed, so users unfamiliar with DomiKnowS can approve it without intervening. Finally, ADS attaches VLM-based learners that enable immediate zero-shot inference, whereas Prompt2DeModel's output is not directly runnable.

Synthesizing NeSy programs from task descriptions necessitates using LLMs as coders. LLMs generate code proficiently for general-purpose languages like Python~\citep{codex, austin2021programsynthesislargelanguage}, but degrade significantly on languages or libraries with scarce training data~\citep{10.1145/3697012,abbassi2025taxonomyinefficienciesllmgeneratedpython}. Consequently, existing NeSy frameworks typically utilize LLMs only with task-specific prompts to extract symbolic representations of natural language queries~\citep{mao2018the,LEFT,nesycoco}, or configure learning components, such as crafting prompts~\citep{VIEIRA}.

ADS also relates to research on end-to-end NeSy program generation. However, prior efforts in this space generally rely on task-specific prompt engineering and custom tool interfaces or executors~\citep{VisProg,ViperGPT,Binding,pan-etal-2023-logic,doi:10.1177/29498732251377499, kamali2025neptuneneuropythonicframeworktunable}. In contrast, we synthesize complete, executable programs for a NeSy library with scarce training data, and our approach is domain-agnostic, supporting any task representable in DomiKnowS.

%% file: Content/Limitations.tex
\section{Limitations}
\label{sec:limitations}

A limitation regarding our test dataset is the risk of pre-training data memorization, which is difficult to entirely rule out. However, our experiments demonstrate that baseline coding assistants failed on these tasks despite having full repository access. Furthermore, ADS successfully resolved three newly authored tasks absent from the pre-training corpus, indicating that its success stems from generalization rather than data recall.

Additionally, our reliance on a proprietary model trades strict reproducibility for the low-latency performance necessitated by our interactive UI. While open-weight alternatives, such as Kimi-K2-Thinking and DeepSeek-R1, offer full reproducibility and achieve higher overall validity, their slower inference speeds currently preclude their use in real-time applications. These open-weight models remain preferable for offline, fully automated settings that do not rely on human-in-the-loop functionality.

%% file: Content/Conclusion.tex
\section{Conclusion}

We introduced AgenticDomiKnowS (ADS), an interactive framework that enables NeSy programming by using natural language task descriptions. By decomposing the workflow into modular stages with self-correcting syntactic and semantic agents, ADS eliminates the steep learning curve typically associated with DomiKnowS. We showed that ADS enables DomiKnowS users and non-users to rapidly construct NeSy programs.

%% file: Content/Appendix/SelectedTasks.tex
\section{DomiKnowS Selected Tasks}
\label{appendix:selectedtasks}

\subsection{Natural Language Processing Tasks}

\subsubsection{Hierarchical News Classification} 

This task involves performing topic classification using the 20 News dataset~\citep{Lang95}, organized into three distinct levels of granularity. The objective is to assign news articles to broad categories at Level 1 (e.g., comp, sci, rec), sub-categories at Level 2 (e.g., windows, crypt, autos), and specific entities at Level 3 (e.g., IBM, hockey, guns). The primary modeling challenge lies in satisfying strict hierarchical consistency constraints, which dictate that a valid classification path must respect parent-child relationships. For example, selecting a class like ``baseball'' necessitates the activation of its parent ``sport'' and ancestor class ``rec''. 

\subsubsection{Spam Classification}
This task utilizes a dataset of emails~\citep{EmailSpam} containing headers, bodies, and spam labels to evaluate consistency modeling. The objective involves deploying two independent classifiers that predict whether a given email is spam or not based on its content. The constraint is a logical consistency requirement between these two models to prevent contradictory outputs. Specifically, the system enforces that the predictions must align such that if the first model predicts ``spam'', the second model is prohibited from predicting ``not spam''.

\subsubsection{Sentiment Analysis}

Using the IMDB dataset~\citep{maas-etal-2011-learning}, this task performs sentiment analysis. One model predicts separate positive and negative probabilities, respectively. A constraint is applied to enforce single-label classification, ensuring every review is categorized as strictly positive or negative.

\subsubsection{Procedural Text Understanding}

This task utilizes the ProPara dataset~\citep{dalvi-etal-2018-tracking} to evaluate a model's ability to track and reason about the dynamic states of entities within procedural text. The objective is to monitor the lifecycle of various elements across a sequence of steps, determining their specific location status (known, unknown, or non-existent) at each stage, while simultaneously identifying the transition actions—such as create, destroy, or other—that drive these changes. The constraints enforce strict causal consistency between actions and state updates. For instance, an action classified as create or destroy must logically correspond to a valid transition between existence and non-existence, ensuring that the predicted sequence of events adheres to the physical laws described in the procedure.

\subsubsection{Causal Reasoning}

This task focuses on reasoning about cause-and-effect relationships within procedural texts using the WIQA dataset~\citep{tandon-etal-2019-wiqa}. The objective is to analyze paragraphs describing sequences of events and determine the qualitative influence between events categorized as: 1) more (positive effect), 2) less (negative effect), or 3) no effect. Logical constraints here include symmetry and transitivity. Symmetric constraints dictate that if the effect of a cause is reversed, e.g., more rain becomes less rain, the effect must invert. Transitivity constraints ensure logical consistency across chains of events. For instance, if event A increases event B, and event B increases event C, then event A must also be inferred to increase event C.

\subsubsection{Belief-Consistent Question Answering}

This task focuses on verifying the truthfulness of facts associated with various entities using the BeliefBank dataset~\citep{kassner-etal-2021-beliefbank}, with a primary emphasis on maintaining global consistency across predictions. The system evaluates a set of candidate sentences describing an entity, aiming to determine which assertions are factually correct. The critical constraint mechanism is derived from a provided relational graph that defines positive and negative correlations between specific attributes. For example, classifying an entity as a ``bird'' might positively imply it ``can fly'', while classifying it as a ``reptile'' might negatively correlate with that same trait. As a result, if a specific attribute is predicted as true, all positively correlated attributes must also be accepted, and all negatively correlated attributes must be rejected, resulting in a non-contradictory set of beliefs for each entity.

\subsubsection{Logical Reasoning Question Answering}

This task evaluates logical reasoning capabilities using the RuleTaker dataset~\citep{RuleTaker}, where the model is presented with a context paragraph containing natural language atomic facts and implication rules. For each context, the model must answer a series of questions verifying the truth value of specific facts. The core constraint is to enforce logical consistency across these predictions. Specifically, if the context establishes a conditional rule (e.g., ``if A then B'') and the model affirms the antecedent, it is structurally required to affirm the consequent, ensuring that the set of predicted answers adheres to the deductive logic.

\subsection{Vision Tasks}

\subsubsection{Hierarchical Image Classification I}

This task applies a hierarchical structure to the standard CIFAR-10 dataset~\citep{CIFAR10}, organizing the ten base classes into two high-level semantic groups: ``animal'' and ``vehicle''. The animal category subsumes natural entities, ``bird'', ``cat'', ``deer'', ``dog'', ``frog'', and ``horse'', whereas the ``vehicle'' category includes artificial objects, ``airplane'', ``automobile'', ``ship'', and ``truck''. The model must respect the strict implication that selecting a specific subclass, e.g., ``ship'', necessitates the activation of its correct parent, ``vehicle''.

\subsubsection{Hierarchical Image Classification II}

This task employs the Animals and Flowers Image Classification dataset~\citep{animalflowers} to evaluate hierarchical classification capabilities. The objective is to correctly identify images at two distinct levels: a coarse-grained superclass level (``animal'' or ``flower'') and a fine-grained subclass level containing specific entities. The constraints require that every image be assigned exactly one specific leaf node, which automatically implies its corresponding parent category.

\subsubsection{Constrained Digit Classification}

This task augments the standard MNIST~\citep{MNISTnormal} handwritten digit recognition challenge by processing images in pairs rather than individually~\citep{DeepProbLog}. The objective is to classify each image into one of ten digit classes. The model is also provided with the ground-truth sum of the two digits in the pair. The defining constraint is arithmetic consistency: the predictions for the two individual images must mathematically add up to the provided sum.

\subsection{Constraint Satisfaction Problem Tasks}

\subsubsection{Sudoku Puzzle}

This task involves solving Sudoku puzzles starting from a partially filled grid. The objective is to populate the remaining cells such that the entire board satisfies strict logical rules. The model represents the puzzle's structural components, explicitly defining concepts for the board, rows, columns, and $3 \times 3$ sub-tables. The constraints strictly enforce the uniqueness property, such that no digit is repeated within any single row, column, or sub-table.

\subsubsection{Eight Queens Puzzle}

This task utilizes a dataset of partially filled chessboards to address the Eight Queens Puzzle. The objective is to determine the correct placement of the remaining queens on an $8 \times 8$ grid to achieve a complete and valid configuration. The constraint for this task requires that no two queens may occupy attacking positions relative to one another.

%% file: Content/Appendix/HumanTest.tex
\section{Instructions for the Human Study}

\label{appendix:humantest}

\subsection*{Overview}
Thank you for participating in this study. The goal of this experiment is to evaluate a system that translates natural language instructions into neuro-symbolic deep learning programs. You will be asked to describe specific tasks to the system, run the generated code, and report on the accuracy and efficiency of the process.

\subsection*{Phase 1: System Access}
To begin the experiment, please open the interface and log in using the credentials provided by the study coordinator.

\subsection*{Phase 2: Task Workflow}
For each of the tasks defined at the end of the instructions, please follow this procedure:

\subsubsection*{Define the Task}
When the system asks you to describe the task, provide a detailed natural language description. Your description should clearly specify:
\begin{itemize}
    \item What the model needs to predict and how the different labels relate to one another.
    \item Explicitly state if there are constraints or if there are none. If multiple constraints exist, define them individually.

\end{itemize}
You do not need to fully define all dataset features upfront. The system will prompt you for the dataset features in subsequent steps.

\subsubsection*{Continue to Navigate the Interface}
\begin{itemize}
    \item Interact with the Interface and provide the requested information until the final code is provided in a Jupyter Notebook.
    \item If you are not familiar with the DomiKnowS framework, strictly skip optional feedback or code editing prompts.
    \item If you make a mistake, click ``New task'' (top-right) to start over. 
\end{itemize}

\subsubsection*{Execute the Program}
Once ADS generates a program:
\begin{enumerate}
    \item Download the generated Jupyter notebook.
    \item Run the notebook locally or upload it to Google Colab (recommended).
    \item Follow any instructions inside the notebook to import the datasets. Minor Python edits may be required to load the data correctly.
\end{enumerate}

\subsection*{Phase 3: Reporting Results}
After attempting all tasks, please compile a report containing:
\begin{itemize}
    \item How many tasks executed successfully using the instances in \texttt{domiknows.datasets}.
    \item Approximate duration for each task in minutes.
    \item Please submit your final notebooks.
\end{itemize}

\subsection*{Task Definitions} 
Please complete the following three tasks:

\subsubsection*{Task (i): Amazon Reviews}
\textbf{Goal:} Predict review ratings (1--5 stars) for an Amazon review dataset. There are no logical constraints between predictions.

\noindent \textbf{Dataset Fields:}
\begin{itemize}
    \item \texttt{review\_id}
    \item \texttt{title}
    \item \texttt{text}
    \item \texttt{label} (0--4, corresponding to 1--5 stars)
\end{itemize}

\subsubsection*{Task (ii): WOS Dataset}
\textbf{Goal:} Predict two labels for each scientific article: a coarse category and a fine category. The predictions should not violate the parent-child relationships.

\noindent \textbf{Hierarchy:}
\begin{itemize}
    \item \textbf{ECE} $\rightarrow$ [Electricity, Digital control, Operational amplifier]
    \item \textbf{Psychology} $\rightarrow$ [Attention, Child abuse, Social cognition, Depression]
    \item \textbf{Biochemistry} $\rightarrow$ [Polymerase chain reaction, Molecular biology, Northern blotting, Immunology]
\end{itemize}

\noindent \textbf{Dataset Fields:}
\begin{itemize}
    \item \texttt{article\_id}, 
    \item \texttt{coarse\_id}
    \item \texttt{fine\_id}
    \item \texttt{text}
    \item \texttt{coarse\_label} (Values 0--2), \item \texttt{fine\_label} (Values 0--10)
\end{itemize}

\subsubsection*{Task (iii): Sequence Labeling of CoNLL Dataset}
\textbf{Goal:} Given a sequence of tokens in a sentence, predict an IOB tag for each token. The tags must be consistent (e.g., an \textbf{``I''} tag cannot appear immediately after an \textbf{``O''} tag).

\noindent \textbf{Dataset Fields:}
\begin{itemize}
    \item \texttt{sentence\_id}
    \item \texttt{token\_ids}
    \item \texttt{tokens}
    \item \texttt{labels} (IOB tags)
\end{itemize}

%% file: Content/Appendix/FullResults.tex
\section{Detailed Evaluation Results}
\label{FullResults}

We evaluate the workflows' performance by targeting both the Knowledge Declaration component and the complete workflow execution. Table~\ref{tab:automaticevaluation} presents the quantitative metrics from the automatic evaluation of the Knowledge Declaration phase, detailing the operational overhead in terms of design iterations, reviewer feedback loops, and syntax corrections required by different models. Complementing these automatic metrics, Table~\ref{tab:tagged} provides the corresponding qualitative manual assessment for these sample runs. Extending the analysis to the end-to-end system, Table~\ref{tab:workflowresults} summarizes the error distribution for the entire workflow under different GPT-5 reasoning levels.

\begin{table*}
    \centering
    \scriptsize
    \setlength{\tabcolsep}{2pt}
    \renewcommand{\arraystretch}{1}

    \begin{tabular}{llcccccccccccc}
    \toprule
    \multirow{2}{*}{\textbf{Model}} & \multirow{2}{*}{\textbf{S\#}} & \multicolumn{12}{c}{\textbf{Tasks}} \\
    \cmidrule(lr){3-14}
     & & \textbf{1} & \textbf{2} & \textbf{3} & \textbf{4} & \textbf{5} & \textbf{6} & \textbf{7} & \textbf{8} & \textbf{9} & \textbf{10} & \textbf{11} & \textbf{12} \\
    \midrule

    \multirow{4}{*}{\textbf{GPT-5 (Minimal)}} 
     & S1  & (2,1,0) & (1,0,0) & (2,1,0) & (1,0,0) & (1,0,0) & (3,2,3) & (3,3,3) & (1,0,0) & (1,0,0) & (3,2,1) & (3,3,0) & (2,1,0) \\[1pt]
     & S2  & (1,0,0) & (1,0,0) & (1,0,0) & (3,3,1) & (1,0,0) & (2,0,1) & (2,1,0) & (3,2,2) & (1,0,0) & (1,0,0) & (3,1,1) & (2,0,1) \\[1pt]
     & S3  & (3,1,1) & (1,0,0) & (3,3,3) & (3,3,1) & (1,0,0) & (3,1,2) & (3,2,0) & (3,0,2) & (1,0,0) & (3,3,0) & (2,1,1) & (2,1,0) \\[1pt]
    \midrule

    \multirow{4}{*}{\textbf{GPT-5 (Low)}} 
     & S1  & (1,0,0) & (3,3,0) & (3,2,0) & (2,0,1) & (1,0,0) & (3,3,2) & (1,0,0) & (3,3,3) & (1,0,0) & (2,1,1) & (3,3,0) & (2,0,1) \\[1pt]
     & S2  & (2,0,1) & (2,1,0) & (3,3,2) & (2,1,1) & (2,0,1) & (3,3,1) & (1,0,0) & (2,1,1) & (2,1,0) & (2,1,0) & (1,0,0) & (2,1,0) \\[1pt]
     & S3  & (2,1,1) & (3,2,0) & (1,0,0) & (2,0,1) & (1,0,0) & (3,3,3) & (1,0,0) & (2,1,1) & (3,3,0) & (2,1,0) & (1,0,0) & (2,1,0) \\[1pt]
    \midrule

    \multirow{4}{*}{\textbf{GPT-5 (Medium)}} 
     & S1  & (3,2,1) & (1,0,0) & (3,2,1) & (3,1,1) & (2,1,1) & (1,0,0) & (1,0,0) & (2,1,1) & (3,3,2) & (3,3,0) & (3,3,0) & (3,2,0) \\[1pt]
     & S2  & (3,2,1) & (1,0,0) & (2,1,0) & (1,0,0) & (2,1,1) & (2,1,0) & (1,0,0) & (3,2,3) & (1,0,0) & (2,1,1) & (3,2,0) & (1,0,0) \\[1pt]
     & S3  & (1,0,0) & (1,0,0) & (3,3,1) & (3,3,1) & (1,0,0) & (2,1,0) & (1,0,0) & (3,2,3) & (3,3,1) & (1,0,0) & (1,0,0) & (1,0,0) \\[1pt]
    \midrule

    \multirow{4}{*}{\textbf{Kimi k2}} 
     & S1  & (3,2,1) & (1,0,0) & (3,2,0) & (1,0,0) & (1,0,0) & (1,0,0) & (3,2,0) & (1,0,0) & (2,1,0) & (2,1,0) & (2,1,0) & (1,0,0) \\[1pt]
     & S2  & (3,2,0) & (3,3,0) & (2,1,0) & (3,3,0) & (1,0,0) & (1,0,0) & (3,2,0) & (2,1,1) & (1,0,0) & (1,0,0) & (1,0,0) & (1,0,0) \\[1pt]
     & S3  & (3,2,0) & (1,0,0) & (1,0,0) & (2,1,0) & (1,0,0) & (1,0,0) & (1,0,0) & (2,1,0) & (2,1,1) & (1,0,0) & (1,0,0) & (2,1,0) \\[1pt]

     \midrule
    \multirow{4}{*}{\textbf{DeepSeek R1}} 
     & S1  & (1,0,0) & (1,0,0) & (3,0,3) & (1,0,0) & (1,0,0) & (1,0,0) & (1,0,0) & (1,0,0) & (1,0,0) & (1,0,0) & (1,0,0) & (1,0,0) \\[1pt]
     & S2  & (1,0,0) & (1,0,0) & (2,1,0) & (1,0,0) & (3,1,1) & (1,0,0) & (1,0,0) & (3,0,3) & (2,0,1) & (1,0,0) & (1,0,0) & (1,0,0) \\[1pt]
     & S3  & (1,0,0) & (1,0,0) & (1,0,0) & (1,0,0) & (1,0,0) & (1,0,0) & (1,0,0) & (1,0,0) & (1,0,0) & (3,0,3) & (1,0,0) & (1,0,0) \\[1pt]

    \bottomrule
    \end{tabular}
    \caption{Automatic evaluation of the Knowledge Declaration section of the workflow across 12 tasks for 3 sample runs (S\#) per LLM models. Each tuple represents the number of attempts made by \textit{Graph Design Agent}, Reviews by the \textit{Graph Reviewer Agent}, and syntax errors caught by the \textit{Graph Execution Agent}.}
    \label{tab:automaticevaluation}
\end{table*}

\begin{table*}
    \centering
    \scriptsize
    \setlength{\tabcolsep}{4pt}
    \renewcommand{\arraystretch}{1.2}

    \begin{tabular}{llcccccccccccc}
    \toprule
    \multirow{2}{*}{\textbf{Model}} & \multirow{2}{*}{\textbf{Sample}} & \multicolumn{12}{c}{\textbf{Tasks}} \\
    \cmidrule(lr){3-14}
     & & \textbf{1} & \textbf{2} & \textbf{3} & \textbf{4} & \textbf{5} & \textbf{6} & \textbf{7} & \textbf{8} & \textbf{9} & \textbf{10} & \textbf{11} & \textbf{12} \\
    \midrule

    \multirow{4}{*}{\textbf{GPT-5 (Minimal)}} 
     & S1  & Se & C & R & C & R & Se & Sy & Se & C & Se & R & C \\[1pt]
     & S2  & C & C & R & Sy & C & Se & R & Se & C & Se & C & C \\[1pt]
     & S3  & R & C & C & C & C & Se & R & Se & C & R & R & C \\[1pt]
    \midrule

    \multirow{4}{*}{\textbf{GPT-5 (Low)}} 
     & S1  & C & R & C & C & C & C & R & Sy & C & R & C & C \\[1pt]
     & S2  & C & C & Sy & C & C & Sy & R & C & C & R & C & C \\[1pt]
     & S3  & C & R & R & C & C & Sy & R & C & Se & R & C & C \\[1pt]
    \midrule

    \multirow{4}{*}{\textbf{GPT-5 (Medium)}} 
     & S1  & C & C & C & C & C & C & C & C & Sy & C & Se & C \\[1pt]
     & S2  & C & R & C & C & C & C & R & Sy & C & R & Se & C \\[1pt]
     & S3  & C & R & C & C & C & C & R & Sy & Se & C & C & C \\[1pt]
    \midrule

    \multirow{4}{*}{\textbf{Kimi k2}}
     & S1  & Sy & C & C & C & C & C & C & C & C & C & C & C \\[1pt]
     & S2  & C & C & C & R & R & C & R & C & C & C & C & C \\[1pt]
     & S3  & C & C & C & C & C & C & R & C & C & C & C & C \\[1pt]

     \midrule
    \multirow{4}{*}{\textbf{DeepSeek R1}} 
     & S1  & C & R & Sy & C & C & C & R & C & C & R & C & C \\[1pt]
     & S2  & C & C & Se & C & C & C & R & Sy & C & R & C & C \\[1pt]
     & S3  & C & R & C & C & C & C & R & C & C & Sy & C & C \\[1pt]

    \bottomrule
    \end{tabular}
    \caption{Manual evaluation of the Knowledge Declaration section of the workflow across 12 tasks for 3 sample runs per LLM models. C (Correct) and R (Redundant) represent the Minimal Valid and Redundant Valid graphs, respectively. The incorrect graphs are shown with Se (Semantic error) or Sy (Syntactic error).}
    \label{tab:tagged}
\end{table*}

\begin{table*}
    \centering
    \scriptsize %
    \setlength{\tabcolsep}{4pt} %
    \renewcommand{\arraystretch}{1.2} %

    \begin{tabular}{llccccccccccccc}
    \toprule
    \multirow{2}{*}{\textbf{Level}} & \multirow{2}{*}{\textbf{Sample}} & \multicolumn{12}{c}{\textbf{Tasks}} & \multirow{2}{*}{\textbf{Total}} \\
    \cmidrule(lr){3-14}
     & & \textbf{1} & \textbf{2} & \textbf{3} & \textbf{4} & \textbf{5} & \textbf{6} & \textbf{7} & \textbf{8} & \textbf{9} & \textbf{10} & \textbf{11} & \textbf{12} & \\
    \midrule

    \multirow{5}{*}{\textbf{Minimal}} 
     & S1 & S & & & & & G & S & G & & G & S & & \multirow{5}{*}{20} \\
     & S2 & & & & S & & G & & S & & & & & \\
     & S3 & S & & & & & S & & & & S & S & & \\
     & S4 & & & & & & S & & & & G & S & & \\
     & S5 & S & & & G & & G & & G & & & & & \\
    \midrule

    \multirow{5}{*}{\textbf{Low}} 
     & S1 & & & & & & S & & G & & S & & & \multirow{5}{*}{14} \\
     & S2 & & & & & & S & & & & S & & & \\
     & S3 & S & & & & & & & & & S & & & \\
     & S4 & & & & G & & G & & & & S & & & \\
     & S5 & S & & & & & S & & S & & S & & & \\
    \midrule

    \multirow{5}{*}{\textbf{Medium}} 
     & S1 & & & & & & & & G & & S & S & & \multirow{5}{*}{11} \\
     & S2 & & & & & & S & & & & & & & \\
     & S3 & G & & & & & & & & & S & & & \\
     & S4 & & & & & & S & & & & & & S & \\
     & S5 & & & & & & G & & & & S & & S & \\
    \midrule

    \bottomrule
    \end{tabular}
    \caption{Evaluation of entire workflow across 12 tasks for 5 sample runs per reasoning level of GPT-5. G and S represent  Graph error and Sensor error, respectively.}
    \label{tab:workflowresults}
\end{table*}

%% file: Content/Appendix/CoderPrompts.tex
\section{Coding Assistant Prompts}
\label{CodingAssistantPrompts}

Figure~\ref{fig:coding_prompt} presents the prompt template used to test our coding assistants during the planning phase. We carefully constructed this prompt with specific instructions to help the model to generate accurate code. For example, we require it to save new scripts in a dedicated test folder and to prevent modifications to the main codebase. Furthermore, we include directives for the assistant to study the provided documentation for correct framework syntax and to automatically execute and self-correct its code until any errors are resolved.

\begin{figure}
    \centering
    \begin{tcolorbox}[colback=gray!5!white, colframe=gray!50!black, arc=2pt, boxrule=0.5pt, left=6pt, right=6pt, top=6pt, bottom=6pt]
    Write a self-contained Python script using the DomiKnowS framework that runs Integer Linear Programming (ILP) inference on the provided dataset with the defined constraints, outputting the final predictions and metrics to the console. Set up the graph and sensors using a distilled GPT-2 model for all learning modules. Ensure that the script runs strictly in inference mode with no training loop. Generate this as a new, separate file in the \texttt{@tests} folder without modifying anything inside the main domiknows repository folder. Use the Python environment \texttt{@<path to Python env>} to run your Python scripts if needed to ensure that it runs correctly. Otherwise, address the errors and rerun the code until the issues are resolved. Refer to the documentation at \texttt{@Docs} to familiarize yourself with the syntax and existing examples.

    \vspace{1em}
    \textbf{Task Description:} \\
    \texttt{<task description>}

    \vspace{1em}
    \textbf{Dataset Example:} \\
    \texttt{<dataset example>}
    \end{tcolorbox}
    \caption{Prompt template provided to the coding assistant during the planning phase.}
    \label{fig:coding_prompt}
\end{figure}